\documentclass[letterpaper, 10pt, conference]{IEEEtran}
\IEEEoverridecommandlockouts
\usepackage{cite}
\usepackage{amsmath,amssymb,amsfonts}
\usepackage{graphicx}
\usepackage{textcomp}
\usepackage{xcolor}
\usepackage{booktabs}
\usepackage{multirow}
\usepackage{array}
\usepackage[letterpaper, left=0.75in, right=0.75in, bottom=0.75in, top=0.75in]{geometry}

\usepackage[ruled,linesnumbered]{algorithm2e}

\def\BibTeX{{\rm B\kern-.05em{\sc i\kern-.025em b}\kern-.08em
    T\kern-.1667em\lower.7ex\hbox{E}\kern-.125emX}}
\begin{document}

\title{\vspace*{0.25in}\LARGE \bf SaFeR: Safety-Critical Scenario Generation for Autonomous Driving Test via Feasibility-Constrained Token Resampling
}

\author{
    {Jinlong Cui, Fenghua Liang, Guo Yang, Chengcheng Tang, Jianxun Cui}
    \thanks{*This research is funded by the Chongqing Natural Science Foundation Innovation and Development Joint Fund (Changan Automobile) (Grant No. CSTB2024NSCQ-LZX0157)
    
    Jinlong Cui, Jianxun Cui are with the School of Traffic and Transportation, Harbin Institute of Technology, Harbin, China and Chongqing Research Institute of HIT, Chongqing, China. Jianxun Cui is the corresponding author. Email: cuijianxun@hit.edu.cn.
    
    Fenghua Liang, Guo Yang, Chengcheng Tang are with the Chongqing Changan Automobile Company Ltd, Chongqing, China.}
}

\maketitle

\begin{abstract}

Safety-critical scenario generation is crucial for evaluating autonomous driving systems. However, existing approaches often struggle to balance three conflicting objectives: adversarial criticality, physical feasibility, and behavioral realism. To bridge this gap, we propose SaFeR: safety-critical scenario generation for autonomous driving test via feasibility-constrained token resampling. We first formulate traffic generation as a discrete next token prediction problem, employing a Transformer-based model as a realism prior to capture naturalistic driving distributions. To capture complex interactions while effectively mitigating attention noise, we propose a novel differential attention mechanism within the realism prior. Building on this prior, SaFeR implements a novel resampling strategy that induces adversarial behaviors within a high-probability trust region to maintain naturalism, while enforcing a feasibility constraint derived from the Largest Feasible Region (LFR). By approximating the LFR via offline reinforcement learning, SaFeR effectively prevents the generation of theoretically inevitable collisions. Closed-loop experiments on the Waymo Open Motion Dataset and nuPlan demonstrate that SaFeR significantly outperforms state-of-the-art baselines, achieving a higher solution rate and superior kinematic realism while maintaining strong adversarial effectiveness.
\end{abstract}


\section{Introduction}
The large-scale deployment of Autonomous Driving Systems (ADS) is contingent upon rigorous safety verification\cite{ding2023survey}. While real-world testing is indispensable, it is inherently limited by cost, efficiency, and safety concerns, especially when evaluating rare yet hazardous interactions such as aggressive cut-in, unprotected turn, or sudden vehicle pull-out. Consequently, simulation-based testing has emerged as a scalable alternative, where the ability to systematically generate realistic and challenging scenarios directly determines the effectiveness of safety assessment.

Recent advances in data-driven scenario generation have significantly improved the realism of simulated traffic behaviors. Large-scale motion datasets have enabled learning-based models to capture complex multi-agent interactions, leveraging paradigms such as diffusion models\cite{chang2025langtraj}, continuous distribution regression\cite{zhou2023query}, and next-token prediction\cite{philion2023trajeglish}\cite{seff2023motionlm}. These approaches excel at modeling the naturalistic distribution of human driving behaviors, producing trajectories that are consistent with real-world data. However, when directly applied to safety testing, such realism-oriented generators tend to underrepresent safety-critical events, as collision-prone interactions are inherently rare in naturalistic datasets.

\begin{figure}
\centering
\includegraphics[width=0.45\textwidth]{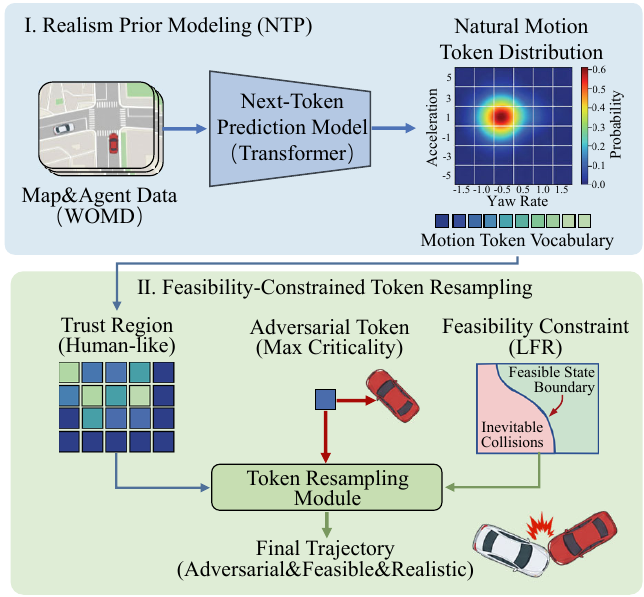}
\caption{Overview of SaFeR. The strategy resamples adversarial tokens from distributions generated by realism prior, constrained by the Largest Feasible Region. }
\label{fig-frontpage}
\end{figure}

To address this imbalance, a growing body of work focuses on safety-critical scenario generation by explicitly inducing adversarial behaviors from background agents. Existing methods optimize collision-related objectives through adversarial trajectory perturbation\cite{wang2021advsim}, gradient-based perturbations\cite{hanselmann2022king}\cite{yin2024regents}, reinforcement learning\cite{zhu2025rl1}, guided diffusion\cite{chang2024safesim}, or reverse generation\cite{liu2025adv}. Despite their effectiveness in increasing collision rates, these approaches often suffer from a fundamental trade-off between \emph{criticality} and \emph{feasibility}. On one hand, unconstrained adversarial optimization can easily produce inevitable or physically implausible collisions, rendering the resulting scenarios invalid for evaluating the decision-making capability of an autonomous vehicle. On the other hand, overly conservative feasibility constraints may significantly limit adversarial strength, leading to scenarios that are feasible but insufficiently challenging.

This tension highlights a core challenge in safety-critical scenario generation: how to synthesize scenarios that are simultaneously \emph{adversarial}, \emph{high-fidelity}, and \emph{feasible}. In particular, a valid safety-critical scenario should (i) induce high collision risk under non-reactive execution, (ii) remain theoretically solvable by a competent ego policy, and (iii) preserve realistic, human-like motion patterns. Achieving all three objectives within a unified framework remains an open problem.

In this paper, we propose SaFeR (\textbf{Sa}fety-critical scenario generation for autonomous driving test via \textbf{Fe}asibility-constrained Token \textbf{R}esampling strategy) to address this challenge. SaFeR tightly integrates a learned realism prior with a feasibility-aware adversarial resample mechanism, as shown in Fig. \ref{fig-frontpage}. Specifically, we formulate scenario generation as a discrete sequence modeling problem and leverage a high-fidelity NTP-based generative model as the realism prior. Crucially, this prior incorporates a tailored Multi-Head Differential Attention (MDA) module that factorizes temporal, agent-agent, and agent-map interactions. By utilizing a paired softmax design, MDA effectively filters out irrelevant background noise, establishing a highly accurate foundation for naturalistic driving behaviors. Building upon this, SaFeR introduces a feasibility constraint based on the Largest Feasible Region (LFR), derived from Hamilton-Jacobi reachability analysis, and approximated via offline reinforcement learning. This constraint explicitly characterizes whether the ego vehicle can theoretically avoid a collision under optimal control, enabling our model to distinguish between challenging-yet-feasible scenarios and inevitable crashes.

The main contributions of this paper are summarized as follows:

\begin{itemize}
    \item We propose \textbf{SaFeR}, a novel framework for safety-critical scenario generation via feasibility-constrained token resampling. By formulating scenario generation as a discrete sequence modeling problem, SaFeR tightly integrates a high-fidelity realism prior with an adversarial resampling strategy. This unified approach effectively resolves the fundamental trade-off between adversarial criticality, physical feasibility, and behavioral realism.
    \item We design a \textbf{multi-head differential attention mechanism} within the realism prior to capture complex spatial-temporal interactions. Utilizing a paired softmax design, the MDA module dynamically filters out irrelevant background attention noise, establishing a highly accurate and naturalistic foundation for human-like driving behaviors.
    \item We introduce a \textbf{feasibility constraint} that leverages offline reinforcement learning to approximate the Largest Feasible Region. This mechanism systematically induces adversarial behaviors within a high-probability trust region while explicitly enforcing a physical feasibility boundary. Consequently, it ensures the generated scenarios are highly challenging yet theoretically solvable, effectively preventing the generation of inevitable collisions.
\end{itemize}


\begin{figure*}
\centering
\includegraphics[width=0.95\textwidth]{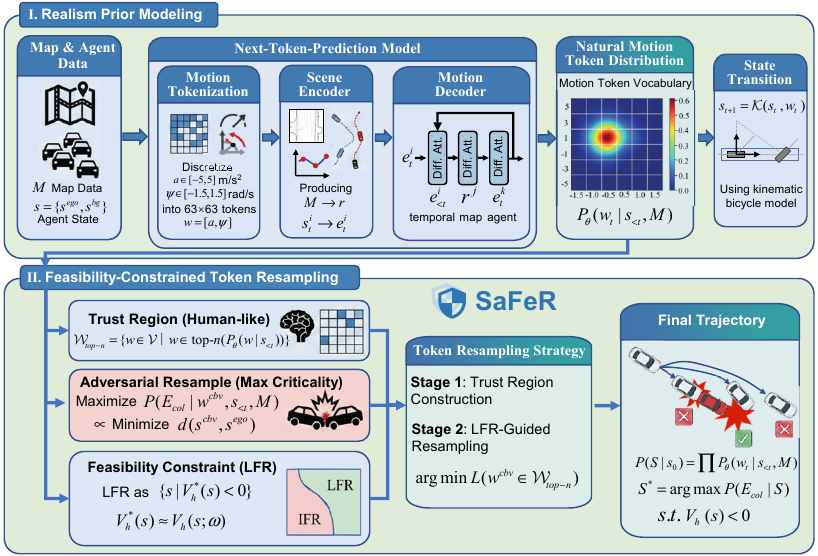}
\caption{Pipeline of SaFeR. The pipeline consists of two core components: (I) Realism Prior Modeling utilizes a Transformer-based NTP model to learn naturalistic motion distributions via motion tokenization and differential attention. (II) Safety-Critical Token Resampling synthesizes the scenario by optimizing for adversarial criticality within a realistic Trust Region, enforced by the Largest Feasible Region constraint to enhance feasibility. }
\label{pipeline}
\end{figure*}

\section{Method}

\subsection{Problem Formulation}\label{problem_formulation}

We define a driving scenario $S$ as a tuple comprising map information $M$ and the temporal evolution of dynamic states for all traffic participants. Let $s_{t} = \{s_{t}^{ego}, s_{t}^{bg}\}$ denote the joint state of the ego vehicle and background agents at time step $t$. A scenario of duration $T$ is represented as: $S = \{M, s_{0:T}\}$.

To model the complex distribution of naturalistic driving behaviors, we reformulate the trajectory generation problem as a discrete sequence modeling task. Instead of predicting continuous states directly, we discretize the action space into a finite vocabulary $\mathcal{V}$ of motion tokens. At each time step $t$, a token $w_t \in \mathcal{V}$ is generated to drive the state transition. Consequently, the generation of a scenario from an initial state $s_0$ is modeled as a product of conditional probabilities:

\begin{equation}
    P(S | s_0) = \prod_{t=1}^{T} P_{\theta}(w_t | s_{<t}, M)
\end{equation}

The objective of safety-critical scenario generation is to find a scenario $S^*$ that maximizes the probability of a collision event $E_{col}$ involving the ego vehicle. 
\begin{equation}\label{eq_op}
    S^* = \mathop{\arg\max}\limits_{S} P(E_{col} | S)
\end{equation}

However, directly optimizing \eqref{eq_op} over the full scenario space is computationally intractable and incompatible with closed-loop simulations that require reactive agent interactions. To overcome these barriers, we approximate the global objective via sequential token-level optimization. This enables greedy selection of the Critical Background Vehicle (CBV) token at each time step. Under this formulation, applying Bayes' theorem, the token selection at time step $t$ reduces to choosing $w_t^{cbv}$ that maximizes the posterior collision probability given the current context:


\begin{equation}  
  \resizebox{0.9\hsize}{!}{$\begin{aligned}
w_t^{cbv*} &= \arg\max_{w^{cbv}_t \in \mathcal{V}} P(w^{cbv}_t | E_{col},s_{<t},M) \\
      &\propto \arg\max_{w^{cbv}_t \in \mathcal{V}} \underbrace{P(E_{col} | w^{cbv}_t, s_{<t},M)}_{\text{Adversarial Criticality}} \cdot \underbrace{P_{\theta}(w^{cbv}_t | s_{<t},M)}_{\text{Realism Prior}}
  \end{aligned}$}
\end{equation}

To solve the optimization objective, we propose a hierarchical framework consisting of two core components:
\begin{itemize}
\item Realism Prior Modeling: We establish a high-fidelity scenario generation model serving as realism prior. This module learns the naturalistic driving distribution $P_{\theta}$ from large-scale datasets, serving as the foundation for realistic behavior generation.
\item Adversarial Optimization: Building upon realism prior, we introduce a token resample strategy. This module utilizes the feasibility constraint to guide the resample process, ensuring that the selected adversarial tokens induce collision risks while maintaining physical feasibility.
\end{itemize}

The overview of our proposed safety critical token resampling  strategy framework is shown in Fig. \ref{pipeline}.

\subsection{Realism Prior Modeling}\label{Sec_Realism}

\textbf{Motion Tokenization.} 
Following \cite{zhao2024kigras}, we adopt acceleration $a$ and yaw rate $\psi$  as motion tokens. We discretize $a\in[-5,5]~\mathrm{m/s^2}$ and $\psi\in [-1.5,1.5]~\mathrm{rad/s}$ into 63 bins separately, yielding a motion vocabulary $V$ of $63\times63$ tokens $w=[a,\psi]$. This fine-grained representation enables flexible modeling of diverse agent actions. The generative model predicts the next token $w \in \mathcal{V}$ for scenario generation.

{\bf Scene Encoder.} The scene context includes map information $M$ and agent information $s$. For agent $i$ at time step $t$, the state $s_t^i=[x_t^i,y_t^i,v_t^i,\theta_t^i,l^i,w^i]$ includes the coordinates, velocity, yaw, length, width. The encoder processes the map information $M$ and agent states $s^i_t$ to produce context embeddings $r$ and agent embeddings $e^i_t$.


{\bf Motion Decoder.} In complex urban driving scenarios, standard Transformer architectures often struggle with attention noise. When modeling dense traffic, the standard scaled dot-product attention tends to indiscriminately assign non-negligible weights to numerous irrelevant background agents or distant map elements, diluting the model's focus on safety-critical interactions. To overcome this limitation, we employ a novel Multi-Head Differential Attention (MDA) mechanism as the core of our motion decoder.

Specifically, we factorize the complex spatial-temporal interaction modeling into three sequential MDA modules: temporal, agent-agent, and agent-map. Taking the $i$-th agent at time step $t$ as an example, we first derive a query from its embedding $e_{t}^{i}$ and apply temporal MDA to attend to its own historical states:

\begin{equation}  
  \resizebox{0.9\hsize}{!}{$\begin{array}{cc}
  e_t^i = MDA\left(q\left(e_t^i\right), k\left(e_{t-\tau}^i, d^{ii}_{t,t-\tau}\right),v\left(e_{t-\tau}^i, d^{ii}_{t,t-\tau}\right)\right) \\[4pt]
0<\tau<t
\end{array}$}
\end{equation}
where $d^{ii}_{t,t-\tau} =\{x^i_t-x^i_{t-\tau}, y^i_t-y^i_{t-\tau}, \theta^i_t-\theta^i_{t-\tau}, \tau\}$ denotes the aggregation of differences in coordinates, yaws, and time at time step $t$ and $t-\tau$ for agent $i$. Subsequently, we employ agent-agent and agent-map differential attention to capture the interaction between agent $i$ and context data:

\begin{equation}  
  \resizebox{0.9\hsize}{!}{$\begin{array}{cc}
e_t^i =MDA\left(q\left(e_t^{i}\right), k\left(e_{t-\tau}^j, d_{t,t-\tau}^{ij}\right), v\left(e_{t-\tau}^j, d_{t,t-\tau}^{ij}\right)\right)  \\[4pt]
 j\in N^i_{agent} ,0\leq\tau<t
\end{array}$}
\end{equation}

\begin{equation}\label{eq_3}
    \begin{array}{cc}
         &e_t^i =MDA\left(q\left(e_t^i\right), k\left(r^j, d^{ij}\right), v\left(r^j, d^{ij}\right)\right)  \\
         & j\in N^i_{map} 
    \end{array}
\end{equation}
where $N^i_{agent}$ and $N^i_{map}$ are the neighborhood set of map and agent tokens determined by a distance threshold.

The critical design of MDA lies in its explicit noise-cancellation property. By utilizing a pair of parallel softmax functions, MDA dynamically subtracts the global attention noise from the primary attention scores. Given input $X \in \mathbb{R}^{N \times d_{input}}$ , we project it into two distinct sets of queries and keys ($Q_1, Q_2, K_1, K_2 \in \mathbb{R}^{N \times d}$) and a shared value $V \in \mathbb{R}^{N \times 2d}$. The output is then computed as the difference between two attention maps:

\begin{equation}  
  \resizebox{0.9\hsize}{!}{$\begin{array}{cc}
[Q_1; Q_2] = q(X) \quad [K_1; K_2] = k(X) \quad V = v(X) \\[6pt]
MDA(X) = \left( \text{softmax}\left(\frac{Q_1 K_1^T}{\sqrt{d}}\right) - \lambda \text{softmax}\left(\frac{Q_2 K_2^T}{\sqrt{d}}\right) \right) V
\end{array}$}
\end{equation}

By subtracting the secondary attention map—which effectively captures the uniform noise profile—modulated by a learnable scalar $\lambda$, the MDA module filters out spurious correlations. In order to synchronize the learning dynamics, we re-parameterize the scalar $\lambda$ as:

\begin{equation} 
\lambda = \exp(\lambda_{q_1} \cdot \lambda_{k_1}) - \exp(\lambda_{q_2} \cdot \lambda_{k_2}) + \lambda_{\text{init}}
\end{equation}
where $\lambda_{q_1}, \lambda_{k_1}, \lambda_{q_2}, \lambda_{k_2} \in \mathbb{R}^d$ are learnable parameters, and $\lambda_{init}\in(0,1)$ is a constant used for the initialization of $\lambda$. Following \cite{ye2024differential}, we use the setting $\lambda_{\text{init}} = 0.8 - 0.6 \times \exp(-0.3 \cdot (l-1))$, where $l\in[1,L]$ represents layer index.

After differential fusion block, we utilize a softmax layer to obtain the probabilities of each token in the motion token vocabulary. 

\begin{equation}\label{eq_6}
    P_{\theta}(w_t^i | s_{<t}, M)=softmax(e_t^i)
\end{equation}

We use the state transition equation to convert the motion token to the next state, formalized as:

\begin{equation}\label{eq_8}
    s_{t+1}^i = \mathcal{K}(s_t^i,w_t^{i})
\end{equation}
where $\mathcal{K}$ represents the kinematic bicycle model.

\subsection{Feasibility Constraint Modeling}\label{Sec_LFR}

Existing adversarial generation methods often pursue high criticality at the cost of feasibility, creating inevitable collisions that are physically impossible for the ego vehicle to avoid. To address this issue, we propose a feasibility constraint which employs the Largest Feasible Region (LFR) to construct a safety boundary. 

We model the feasibility constraint using Hamilton-Jacobi Reachability (HJR) analysis\cite{bansal2017hamilton}.  Let $h(s^{}_t)$ be a constraint violation function representing the immediate safety status. We adopt a sparse indicator form to facilitate stable training. 

\begin{equation}\label{eq_12}
h(s^{}_t) =
\begin{cases}
-1, &   d(s^{ego}_t,s^{cbv}_t) > d_{th}  \\
M,  &   d(s^{ego}_t,s^{cbv}_t) \le d_{th} 
\end{cases}
\end{equation}

where $d(\cdot,\cdot)$ represents the minimum distance between the bounding boxes. $d_{th}$ is a safety distance threshold. $M$ is a large penalty constant.

According to \cite{bansal2017hamilton}, the optimal feasible state-value function $V^*_h(s^{})$ and the optimal feasible action-value function $Q_h^*(s^{},a^{})$ are defined as the minimum of the maximum future constraint violation under an optimal driving policy $\pi^{}$:
\begin{equation}\label{eq_13}
\begin{aligned}
   V^*_h(s^{}) := \min_{\pi^{}} \max_{t \in [0, T]} h(s_t^{}) \\
s_0^{}=s^{},a_t^{}\sim \pi^{}(\cdot | s_t^{}) 
\end{aligned}
\end{equation}

\begin{equation}\label{eq_14}
\begin{aligned}
&Q^*_h(s^{},a^{}) := \min_{\pi^{}} \max_{t \in [0, T]} h(s_t^{})\\
s_0^{}&=s^{},a_0^{}=a^{},a^{}_{t+1}\sim \pi^{}(\cdot | s_{t+1}^{})
\end{aligned}
\end{equation}

Mathematically, the zero-sublevel set of this function:
\begin{equation}\label{eq_15}
\mathcal{S}^*_f = \{s^{} \mid V^*_h(s^{}) \le 0\}
\end{equation}
where $\mathcal{S}^*_f$ represents the Largest Feasible Region (LFR)—the set of states from which the ego vehicle can theoretically avoid a collision.

Although LFR enforces effective constraints, calculating the optimal feasible value function based on \eqref{eq_13}\eqref{eq_14} requires Monte-Carlo estimation via interacting with the environment. In scenario generation, however, the ego vehicle typically follows a fixed policy during such interactions. As a result, the feasible region inferred from these interactions only constitutes a subset of the LFR. 

To solve this problem, we use the method proposed in \cite{chen2024frea} which approximates $V^*_h$ using a neural network. To decouple the policy from the value estimation and handle the distribution shift in offline data, we employ an expectile regression approach\cite{kostrikov2021offline}, simultaneously learning the feasibility action-value function $Q_h(s, a; \phi)$ and the state-value function $V_h(s; \omega)$ by minimizing the following functions:

\begin{equation}\label{eq_16}
\begin{aligned}
\mathcal{L}_{V_h}(\omega) &= \mathbb{E}_{(s,a) \sim \mathcal{D}} \left[ L_{rev}^\tau \left( Q_h(s, a; \phi) - V_h(s; \omega) \right) \right]\\[4pt]
&L_{rev}^\tau(u) = |\tau - \mathbb{I}(u>0)| u^2
\end{aligned}
\end{equation}

\begin{equation}\label{eq_17}
\begin{aligned}
\mathcal{L}_{Q_h}(\phi) = \mathbb{E}_{(s,a,s') \sim \mathcal{D}} \left[ \left( \hat{Q} - Q_h(s, a; \phi) \right)^2 \right]\\
\hat{Q} = (1-\gamma)h(s) + \gamma \max \{ h(s), V_h(s'; \omega) \}
\end{aligned}
\end{equation}
where $\tau$ is a hyperparameter balancing the estimation. $\gamma$ is a discount factor. 

Once trained, the network $V_h(s; \omega)$ is frozen and used as an approximation of $V_h$ in \eqref{eq_10} during token resampling.

\subsection{Safety-Critical Token Resampling Strategy}\label{Sec_TR}

Leveraging the pre-trained feasibility value function $V_h$ and realism prior model $P_\theta$, we devise a safety-critical token resampling strategy to resample adversarial action $w_t^{cbv}$ for the critical background vehicle. This process is formulated as a Two-Stage Constrained Search:

Stage 1: Trust Region Construction. 
To ensure the adversarial vehicle behaves like a human driver, we limit the search space to the high-probability manifold of the realism prior model. We define the trust region $\mathcal{W}_{top-n}$ as the top-$n$ most probable tokens predicted by the realism prior:

\begin{equation}
    \mathcal{W}_{top-n} = \{ w \in \mathcal{V} \mid w \in \text{top-}n(P_{\theta}(w | s_{<t},M)) \}
\end{equation}

This truncation guarantees that the selected token is inherently realistic.

Stage 2: LFR-Guided Token Resampling.
Within the trust region $\mathcal{W}_{top-n}$, we seek a token $w^*$ that maximizes adversarial criticality without violating the feasibility constraint. We define an adversarial loss, $\mathcal{L}(w)$, which dynamically switches objectives based on the safety status of ego vehicle.

\begin{equation}\label{eq_10}
\begin{aligned}
\mathcal{L}(w_t^{cbv})=
\begin{cases}
d(s_{t+1}^{cbv},s^{ego}_{t+1}),
& V_h(s_{t+1}^*)\le 0, \\[4pt]
V_h(s_{t+1}^*)+\Delta,
& V_h(s_{t+1}^*)>0,
\end{cases}
\\[4pt]
s_{t+1}^{cbv}=\mathcal{K}(s_{t}^{cbv},w_t^{cbv}) \quad s_{t+1}^*=\{s_{t+1}^{ego},s_{t+1}^{cbv}\}\quad \quad
\end{aligned}
\end{equation}
where $\mathcal{K}$ is defined in \eqref{eq_8}, $\Delta$ is a large constant margin added to ensure that any infeasible token always has a higher loss than any feasible token. This loss function enforces a hierarchical priority mechanism:
\begin{itemize}
    \item Feasible Region, $V_{h}^{*} \le 0$: If the ego vehicle is within the LFR, the strategy acts greedily to increase criticality by minimizing the euclidean distance between vehicles.
    \item Infeasible Region, $V_{h}^{*} > 0$: If a candidate token pushes the ego vehicle outside the LFR (i.e., towards an unavoidable crash), the loss heavily penalizes this violation. The optimization objective shifts from distance minimization to feasibility recovery (minimizing $V_h^*$).
\end{itemize}

Finally, the optimal adversarial token is obtained by:

\begin{equation}
    w_{t}^{cbv*} = \arg\min_{w \in \mathcal{W}_{top-n}^{cbv}} \mathcal{L}(w_t^{cbv})
\end{equation}

This strategy ensures that the generated scenarios are adversarial (via distance minimization), high fidelity(via realism constraint), and feasible (via LFR constraint), avoiding the generation of inevitable accidents.

\section{Experiments}

\subsection{Experimental Setup}
\label{setup}

\textbf{Dataset.} We train our NTP-based scenario generation model on Waymo Open Motion Dataset(WOMD)\cite{waymo} and evaluate on both WOMD and nuPlan\cite{caesar2021nuplan} to verify the effectiveness of the realism prior.

For LFR offline reinforcement learning, we collect interaction data under multiple planner configurations to improve coverage of diverse conflict patterns. Specifically, we use SMART \cite{wu2024smart} and DiffusionPlanner \cite{zheng2025difplan} as ego policies, while our NTP-based scenario generation model (mentioned in Section~\ref{Sec_Realism}) controls background agents, collecting 100k interaction instances for each configuration. To further enrich highly adversarial and conflict-prone scenarios, we additionally employ SafeSim \cite{chang2024safesim} as the critical background vehicle and SMART as the ego policy to generate another 100k interaction instances. In total, we obtain 300k interaction samples to train the LFR feasibility value network, ensuring broad coverage of both naturalistic and safety-critical traffic configurations.

We import 1000 scenarios involving complex vehicle interactions from the WOMD and 1000 from nuPlan as the dataset to evaluate the safety-critical scenario generation model. Each scenario in the dataset contains a traffic participant labeled as object of interest regarding the ego vehicle, which is also designated as the critical background agent in our experiments.

\textbf{Implementation.} All the experiments are conducted in a closed-loop manner within Waymax\cite{gulino2023waymax}, a data-driven AD simulator. The hyperparameters are listed in TABLE \ref{tab:hyperparameters}.

\begin{table}[h]
    \centering
    \caption{Hyperparameter Settings}
    \label{tab:hyperparameters}
    \renewcommand{\arraystretch}{1.2} 
    \begin{tabular}{l|c|c}
        \hline
        \textbf{Parameter} & \textbf{Symbol} & \textbf{Value} \\
        \hline
        \hline
        \multicolumn{3}{l}{\textit{Scenario Generation Model}} \\
        \hline
        Attention layers & $L$ & 3 \\
        Attention Heads & $H$ & 8 \\
        \hline
        \multicolumn{3}{l}{\textit{Feasibility Value Network}} \\
        \hline
        Discount Factor & $\gamma$ & 0.99 \\
        Expectile Parameter & $\tau$ & 0.8 \\
        Safety Distance Threshold & $d_{th}$ & 0.3 \\
        penalty constant & $M$ & 16 \\
        \hline
        \multicolumn{3}{l}{\textit{Token Resampling Strategy}} \\
        \hline
        Trust Region Size & $n$ & 20 \\
        Penalty Margin & $\Delta$ & 50.0 \\
        Planning Frequency & $f$ & 2 Hz \\
        \hline
    \end{tabular}
\end{table}

\subsection{Realism Evaluation}

\begin{table}[t]
    \centering
    \small
    \caption{General Scenario Quantitative results on WOMD and nuPlan.}
    \label{tab-real}
    \scalebox{0.95}{
    \begin{tabular}{l l l l l l}
    \toprule
         Dataset & Method & Real.$\uparrow$ & Kin.$\uparrow$ & Inter.$\uparrow$ & Map.$\uparrow$ \\
        \midrule
        \multirow{6}{*}{WOMD}
         & Diffusion\cite{zheng2025difplan} &0.7644&0.4771&0.8135& 0.8653\\
         & QCNet\cite{zhou2023query} & 0.7013 & 0.4298& 0.7150& 0.8387\\
         & GUMP\cite{hu2024gump}& 0.7446 & 0.4737 & 0.7791 & 0.8551\\
         & SMART\cite{wu2024smart}& 0.7595 & 0.4749 & 0.8039 & 0.8652\\
         & \textbf{SaFeR(ours)} & \textbf{0.7730} & \textbf{0.4798} & \textbf{0.8254} & \textbf{0.8731}\\
        \midrule
        \multirow{6}{*}{nuPlan}
         & Diffusion\cite{zheng2025difplan} &0.6597&0.4151&0.7417& 0.6939\\
         & QCNet\cite{zhou2023query} & 0.5840 & 0.3796& 0.6510& 0.6146\\
         & GUMP\cite{hu2024gump}& 0.5674 & 0.3299 & 0.6341 & 0.6173\\
         & SMART\cite{wu2024smart}& 0.6453 & 0.4114 & 0.7420 & 0.6546\\
         & \textbf{SaFeR(ours)} & \textbf{0.6760} & \textbf{0.4197} & \textbf{0.7512} & \textbf{0.7257}\\
        \bottomrule 
    \end{tabular}}
\end{table}

We evaluate our NTP-based scenario generation model against a broad spectrum of open source approaches, covering fundamentally different modeling paradigms, including diffusion models\cite{zheng2025difplan}, continuous distribution regression models \cite{zhou2023query}\cite{hu2024gump}, and next token prediction models\cite{wu2024smart}. 

The metrics from Waymo Sim Agent Challenge 2025 \cite{montali2023waymoc} include kinematic, interactive, and map-based metrics, which are aggregated into a realism meta metric. As shown in TABLE \ref{tab-real}, our model outperforms all baselines in terms of the realism meta-metric.

Notably, the incorporation of the Multi-Head Differential Attention (MDA) mechanism drives the most pronounced improvements in the Interactive and Map-based metrics. These results demonstrate that MDA empowers the model to effectively isolate critical interaction cues from dense environments. Ultimately, this superior fidelity serves as a strong realism prior for the safety-critical token resampling strategy.

\begin{figure}
\centering
\includegraphics[width=0.45\textwidth]{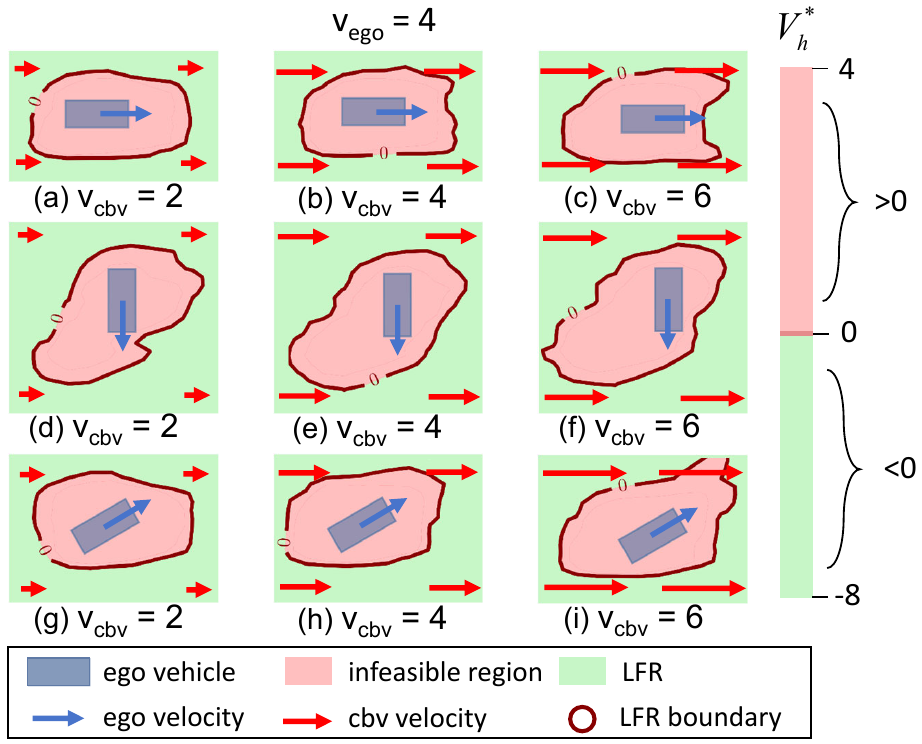}
\caption{Visualization of the learned LFR of ego vehicle. The red area depicts the Infeasible Region ($V_h^* > 0$), representing the set of spatial configurations for the critical background vehicle that lead to unavoidable collisions given the specified velocity vectors. The green area denotes the Largest Feasible Region (LFR) ($V_h^* \le 0$), where a feasible evasion policy theoretically exists for the ego vehicle. }
\label{fig-LFR}
\end{figure}

\subsection{Feasibility Evaluation}

To validate the physical consistency of the learned feasibility value function $V_h(s)$, we visualize the decision boundaries of the Largest Feasible Region (LFR) under varying kinematic configurations.
As shown in Fig. \ref{fig-LFR}a-c, we observe a monotonic expansion of the infeasible region as  $v^{cbv}$ increases from $2 m/s$ to $6 m/s$. Higher approaching speeds significantly reduce the temporal window for reaction and the spatial envelope for evasion maneuvers. Consequently, the set of states from which a collision is mathematically inevitable expands. 

As depicted in Fig. \ref{fig-LFR}d-e, the morphology of the infeasible region exhibits strong anisotropy, rotating and deforming dynamically in alignment with the velocity vectors of the interacting agents. The boundary extends further along the longitudinal direction of travel, reflecting the braking distance and the non-holonomic constraints of the vehicle (i.e., cars cannot move sideways instantly).

These visualizations confirm that our offline-learned $V_h(s)$ effectively captures the complex, non-linear relationship between vehicle dynamics and collision risks. It provides a physically grounded metric that allows the resampling strategy to distinguish between imminent danger (which should be penalized) and challenging but safe interactions (which should be encouraged).


\begin{figure*}
\centering
\includegraphics[width=1\textwidth]{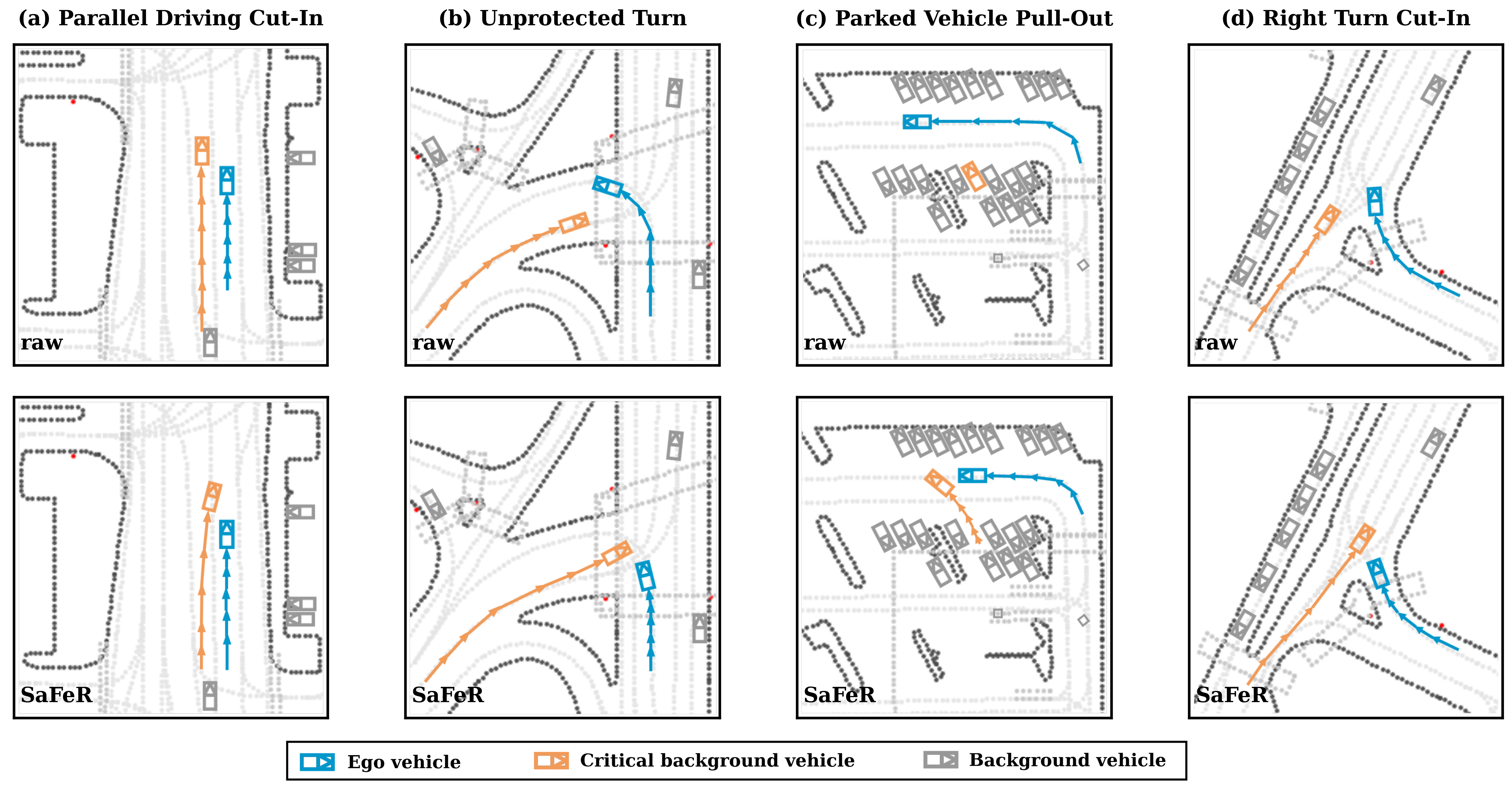}
\caption{Qualitative comparison between the raw scenario and safety-critical scenario generated by SaFeR. The critical background vehicle is controlled by SaFeR, while the ego vehicle is controlled by DiffusionPlanner\cite{zheng2025difplan}.  }
\label{fig_safetycritical}
\end{figure*}

\subsection{Criticality Evaluation}

\textbf{Qualitative Experiments.} To intuitively demonstrate the effectiveness of SaFeR, we present a qualitative comparison between the raw scenarios and the safety-critical scenarios generated by SaFeR in Fig. \ref{fig_safetycritical}. The comparison covers four representative urban driving cases: parallel driving, unprotected intersection navigation, parking lot interactions, and turning maneuvers.

In the parallel driving scenario(Fig. \ref{fig_safetycritical}a), the CBV originally maintains its lane. SaFeR generates a sharp cut-in trajectory, minimizing the lateral separation and forcing the ego vehicle to slow down.

In the intersection scenarios(Fig. \ref{fig_safetycritical}b$\And$d), the CBV refuses to yield. It aggressively competes for the right-of-way by cutting across the ego vehicle's predicted path, simulating common real-world accidents caused by misjudgment or aggression.

In the Parked Vehicle Pull-Out case(Fig. \ref{fig_safetycritical}c), SaFeR identifies a high-risk opportunity where a stationary vehicle suddenly merges into the lane just as the ego vehicle approaches.

Crucially, while the generated behaviors are adversarial, the trajectories remain smooth and kinematically feasible. This confirms the effectiveness of the Trust-Region Constraint, which ensures the adversarial tokens are sampled from the high-probability manifold of the base generation model, preserving the naturalism of human driving behaviors while maximizing criticality.


\begin{table}[t]
    \centering
    \small
    \caption{Quantitative comparison of safety-critical scenario generation performance. Results are the average of 10 runs with varied seeds.}
    \label{tab-quantitative_results}
    \scalebox{0.95}{
    \begin{tabular}{l l l l l l}
    \toprule
         Dataset & Method & CR $\uparrow$ & SR $\uparrow$ & VJ $\downarrow$ & AJ $\downarrow$ \\
        \midrule
        \multirow{6}{*}{WOMD}
        &ReGentS \cite{yin2024regents}  & 0.733 & 0.510 & 0.259 & 0.614 \\
        &FREA \cite{chen2024frea} & 0.536 & 0.719 & 0.213 & 0.566 \\
        &SAFE-SIM \cite{chang2024safesim} & 0.616 & 0.549 & 0.197 & 0.508 \\
        &ADV-BMT \cite{liu2025adv} & \textbf{0.915} & 0.324 & 0.251 & 0.603 \\
        &\textbf{SaFeR(ours)} & 0.761 & \textbf{0.865} & \textbf{0.161} & \textbf{0.499} \\
        \midrule
        \multirow{6}{*}{nuPlan}
        &ReGentS \cite{yin2024regents}  & 0.617 & 0.534 & 0.291 & 0.627 \\
        &FREA \cite{chen2024frea} & 0.590 & 0.691 & 0.225 & 0.567 \\
        &SAFE-SIM \cite{chang2024safesim} & 0.749 & 0.615 & 0.201 & 0.519 \\
        &ADV-BMT \cite{liu2025adv} & \textbf{0.877} & 0.310 & 0.287 & 0.610 \\
        &\textbf{SaFeR(ours)} & 0.757 & \textbf{0.801} & \textbf{0.179} & \textbf{0.510} \\
        \bottomrule
    \end{tabular}}
\end{table}

\textbf{Quantitative Experiments.} We compare SaFeR with state-of-the-art safety-critical scenario generation model including: gradient-based optimization method\cite{yin2024regents}, feasibility-guided method\cite{chen2024frea}, reverse generation method\cite{liu2025adv}, diffusion-based model\cite{chang2024safesim}.

To comprehensively evaluate the generated scenarios, we establish a Dual-Stage Evaluation Protocol\cite{rempe2022Strive}. In the first phase, the ego vehicle follows a non-reactive Log Replay trajectory to measure the Collision Rate (CR), where a higher CR indicates that the generated Critical Background Vehicle (CBV) effectively creates valid adversarial conflicts. In the second phase, a reactive DiffusionPlanner\cite{zheng2025difplan} controls the ego vehicle to assess the Solution Rate (SR); a high SR confirms that the generated scenarios remain feasible and do not force the ego into inevitable accidents. Additionally, we compute the Jensen-Shannon Divergence of velocity (VJ) and acceleration (AJ) between the generated CBV and the ground truth to ensure the adversarial behaviors remain kinematically realistic and consistent with naturalistic driving distributions.

TABLE \ref{tab-quantitative_results} presents the quantitative comparison. While ADV-BMT achieves the highest CR, it exhibits a disproportionately low SR, indicating that the majority of generated collisions are theoretically unavoidable and thus invalid for safety tests. In contrast, SaFeR achieves the highest SR, demonstrating that our LFR constraint effectively filters out infeasible scenarios. Furthermore, SaFeR outperforms all baselines in terms of realism, achieving the lowest VJ and AJ. This confirms that by incorporating the trust region, our framework successfully generates valid safety-critical scenarios that strike a balance between high adversarial criticality and strict adherence to human-like kinematic distributions.

\subsection{Ablation Studies}

\textbf{Multi-Head Differential Attention(MDA) and Largest Feasible Region(LFR).} To validate the individual contributions of the MDA and the LFR constraint, we conducted a component-wise ablation study. As shown in TABLE \ref{tab:ablation_mda_lfr}, removing the LFR constraint (w/o LFR) yields the highest Collision Rate (0.827) but causes a severe drop in the Solution Rate (0.527), indicating that unconstrained adversarial optimization generates theoretically unavoidable and physically invalid collisions. Conversely, removing the differential attention module (w/o MDA) degrades both the solution rate (0.720) and kinematic realism (VJ of 0.205, AJ of 0.519), demonstrating that attention noise compromises the high-fidelity naturalistic foundation required for valid resampling. Ultimately, the complete SaFeR framework successfully harmonizes these components, achieving the highest SR (0.865) and superior kinematic fidelity (lowest VJ of 0.161 and AJ of 0.499) while maintaining a robust adversarial criticality (CR of 0.761). This confirms that both the MDA mechanism and the LFR constraint are indispensable for synthesizing safety-critical scenarios that are simultaneously challenging, feasible, and realistic.

\begin{table}[htbp]
\centering
\caption{Ablation Study on the Multi-Head Differential Attention(MDA) and Largest Feasible Region(LFR). All experiments are based on the WOMD dataset.}
\label{tab:ablation_mda_lfr}
\begin{tabular}{lcc|cccc}
\toprule
Setting & MDA & LFR & CR$\uparrow$ & SR$\uparrow$ & VJ$\downarrow$ & AJ$\downarrow$ \\
\midrule
w/o MDA & - & \checkmark & 0.793 & 0.720 & 0.205 & 0.519 \\
w/o LFR & \checkmark & - & \textbf{0.827} & 0.527 & 0.187 & 0.501 \\
SaFeR & \checkmark & \checkmark & 0.761 & \textbf{0.865} & \textbf{0.161} & \textbf{0.499} \\
\bottomrule
\end{tabular}
\end{table}

\begin{figure}
\centering
\includegraphics[width=0.45\textwidth]{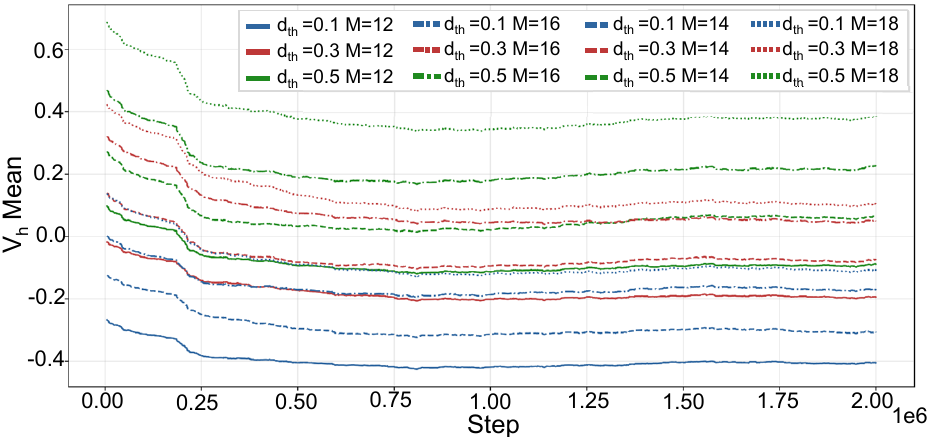}
\caption{The mean of Feasible Value Network $V_h$ during the training with multiple combinations of penalty constant $M$ and safety distance threshold $d_{th}$. }
\label{fig-Vhtraining}
\end{figure}

\textbf{Penalty constant $M$ and Safety distance threshold $d_{th}$.} We tracked the convergence of the mean $V_h$ under various combinations of the penalty constant $M$ and safety distance threshold $d_{th}$, as shown in Fig. \ref{fig-Vhtraining}. According to \cite{zheng2024safe}, optimal performance is achieved when the mean of $V_h$ is close to zero. Therefore, we selected $d_{th}=0.3$ and $M=16$, where the mean of $V_h$ stabilizes closest to 0, ensuring that the learned $V_h$ provides the most accurate approximation of the true feasibility boundary without biasing towards overly conservative or aggressive estimations.

\begin{table}
    \centering
    \caption{Ablation study on the Trust Region size ($n$). The setting used in our final method is marked with *.}
    \label{tab:ablation_top_n}
    \renewcommand{\arraystretch}{1.2}
    \setlength{\tabcolsep}{10pt} 
    \begin{tabular}{c|cccc}
        \toprule
        \textbf{Trust Region ($n$)} & \textbf{CR} ($\uparrow$) & \textbf{SR} ($\uparrow$) & \textbf{VJ} ($\downarrow$) & \textbf{AJ} ($\downarrow$) \\
        \midrule
        10 & 0.203 & 0.901 & 0.153 & 0.471 \\
        20$^*$ & 0.761 & 0.865 & 0.161 & 0.499 \\
        50 & 0.794 & 0.817 & 0.495 & 0.630 \\
        100 & 0.805 & 0.801 & 0.517 & 0.796 \\
        \bottomrule
    \end{tabular}
\end{table}

\textbf{Trust Region Size}: The trust region size $n$ controls the trade-off between adversarial capability and behavioral realism. As presented in Table \ref{tab:ablation_top_n}, increasing $n$ from 10 to 20 results in a substantial improvement in CR, indicating that a broader search space is necessary to identify critical trajectories. However, further increasing $n$ to 50 or 100 yields marginal gains in SR but causes a sharp degradation in realism. Consequently, we select $n=20$ as the optimal setting.

\section{Conclusion}

In this paper, we proposed SaFeR: safety-critical scenario generation for autonomous driving test via feasibility-constrained token resampling strategy. To establish a robust generation foundation, we first introduced a novel Multi-Head Differential Attention mechanism that effectively mitigates attention noise, yielding a highly naturalistic realism prior for complex spatio-temporal interactions. By harmonizing this high-fidelity realism prior with an explicit Largest Feasible Region constraint, SaFeR effectively resolves the conflict between adversarial criticality and physical feasibility. Specifically, our two-stage resampling strategy induces challenging interactions within a realistic trust region while preventing theoretically inevitable collisions. Closed-loop experiments on the WOMD and nuPlan demonstrate that SaFeR significantly outperforms state-of-the-art baselines in both solution rate and kinematic realism, without compromising adversarial strength. These results highlight the necessity of incorporating token-level feasibility constraints and provide an effective method for evaluating the safety of Autonomous Driving Systems.


\bibliographystyle{IEEEtran}
\bibliography{refs}

\end{document}